%% file: main.tex
\documentclass{article} 
\usepackage{collas2022_conference,times}

\input{math_commands.tex}

\usepackage{hyperref}
\hypersetup{
    colorlinks=true,
    linkcolor=red,
    filecolor=magenta,      
    urlcolor=blue,
    citecolor=purple,
    pdftitle={Overleaf Example},
    pdfpagemode=FullScreen,
    }

\usepackage{url}
\usepackage{booktabs}
\usepackage{arydshln}
\usepackage{caption}
\usepackage{subcaption}
\usepackage{amsmath}
\usepackage{mathtools}
\usepackage{wrapfig}

\newcommand{\as}[1]{\renewcommand{\arraystretch}{#1}}
\newcommand{\param}{\theta}
\newcommand{\mparam}{\eta}

\newcommand{\loss}{\mathcal{L}}

\newcommand{\inne}{(\text{inner})}
\newcommand{\oute}{(\text{outer})}
\newcommand{\ent}{\text{ent}}
\newcommand{\mg}{\text{MG}}
\newcommand{\bmg}{\text{BMG}}

\title{Meta-Gradients in Non-Stationary Environments}

\author{Jelena Luketina\thanks{Work done during an internship at DeepMind. Contact: \texttt{jelena.luketina@cs.ox.ac.uk}} \\ 
University of Oxford
\And 
Sebastian Flennerhag \\
DeepMind
\And 
Yannick Schroecker \\
DeepMind
\AND 
David Abel \\
DeepMind
\And 
Tom Zahavy \\
DeepMind
\And 
Satinder Singh \\
DeepMind
}

\collasfinalcopy 

\begin{document}

\maketitle

\input{core/abstract}
\input{core/introduction}

\input{core/method}

\input{core/related_work}

\input{core/experiments}

\section{Conclusions}

We studied the performance and properties of white-box meta-gradients in non-stationary environments. To study the effect of adding contextual information to the learned optimizer, we focused on formulations of meta-gradients where the learned meta-parameter values are functions of selected context features. We found that adding more contexual information is almost always beneficial for lifetime performance. Inspection of learned meta-parameter schedules and functions provides evidence of faster adaptation for meta-gradients with contextual information and convergence of meta-parameter functions over training. 
When looking at the effect of increasing the rate of non-stationarity, we find that the meta-gradients without context, in contrast to meta-gradients with context, do not offer a large consistent advantage over fixed meta-parameter schedules. 
An interesting avenue for future research are studies of contextual meta-gradients in continual supervised learning setting and in non-stationary environments with less repeatability of context features.

\section*{Acknowledgements}

We would like to thank Junhyuk Oh, Risto Vuorio, Shimon Whiteson and anonymous reviewers for useful feedback on the earlier versions of this paper. 
Our work was funded by DeepMind.

\bibliography{main}
\bibliographystyle{collas2022_conference}

\appendix
\input{core/appendix}

\end{document}

%% file: math_commands.tex
\usepackage{amsmath,amsfonts,bm}

\def\eqref#1{equation~\ref{#1}}

\def\1{\bm{1}}

\DeclareMathAlphabet{\mathsfit}{\encodingdefault}{\sfdefault}{m}{sl}
\SetMathAlphabet{\mathsfit}{bold}{\encodingdefault}{\sfdefault}{bx}{n}

\DeclareMathOperator*{\argmax}{arg\,max}

%% file: core/abstract.tex
\begin{abstract}
Meta-gradient methods \citep{Xu:2018metagradient, Zahavy:2020se} offer a promising solution to the problem of hyperparameter selection and adaptation in non-stationary reinforcement learning problems. However, the properties of meta-gradients in such environments have not been systematically studied. 
In this work, we bring new clarity to meta-gradients in non-stationary environments. Concretely, we ask: (i) how much information should be given to the learned optimizers, so as to enable faster adaptation and generalization over a lifetime, (ii) what meta-optimizer functions are learned in this process, and (iii) whether meta-gradient methods provide a bigger advantage in highly non-stationary environments.
To study the effect of information provided to the meta-optimizer, as in recent works \citep{Flennerhag:2021bml, Almeida:2021}, we replace the tuned meta-parameters of fixed update rules with learned meta-parameter functions of selected context features. The context features carry information about agent performance and changes in the environment and hence can inform learned meta-parameter schedules.
We find that adding more contextual information is generally beneficial, leading to faster adaptation of meta-parameter values and increased performance. We support these results with a qualitative analysis of resulting meta-parameter schedules and learned functions of context features. Lastly, we find that without context, meta-gradients do not provide a consistent advantage over the baseline in highly non-stationary environments. 
Our findings suggest that contextualising meta-gradients can play a pivotal role in extracting high performance from meta-gradients in non-stationary settings.
\end{abstract}


%% file: core/introduction.tex
\section{Introduction} \label{sec:intro}

Meta-gradient approaches to learning adaptive optimizers are a promising complement gradient-based optimizers in reinforcement learning (RL). By adapting relevant optimization hyperparameters or the entire update rule to the current domain, they often outperform well-tuned gradient-based optimizers \citep{Schraudolph:1999meta, mahmood2012tuning, Andrychowicz:2016tf, Xu:2018metagradient}. 
The adaptability of meta-learned optimizers is particularly relevant for non-stationary environments: as the agent continues learning during its lifetime, appropriate hyperparameter values for an optimizer are likely to change over time and can be impossible to determine in advance \citep{parker2022automated}. Despite this promise, the performance and properties of meta-gradients in the context of reinforcement learning in non-stationary environments have not been systematically studied, which is the focus of our work.
The question we focus on is how does the information provided to the meta-optimizer as well as the rate of environment non-stationarity, effect the performance and properties  
of meta-gradients. 

Most of the prior work in this area can be classified within the framework of white- and black-box optimization (see Related Work in \cite{xu2020meta} for a more detailed classification of prior work). Black-box methods express the entire update rule as a rich parametric function, typically a recurrent neural network,
and learn the parameters of this function in an end-to-end manner \citep{xu2020meta, Oh:2020discoveringrl, Kirsch2021IntroducingST}.
In contrast, white-box methods tune the hyperparameters of the optimization algorithm 
\citep{mahmood2012tuning, Xu:2018metagradient, Zahavy:2020se}. 
We limit the scope of our analysis to white-box meta-gradient methods for self-tuning  hyperparameters, which have to date shown the greatest empirical gains in RL \citep{Xu:2018metagradient, Zahavy:2020se, Flennerhag:2021bml}.

Since white-box methods are almost completely memory-less and only tune several hyperparameters based on their local influence on agent performance, standard white-box meta-gradients are only capable of tracking good solutions \citep{Sutton:2007tracking} and lack the ability to learn and benefit from past experience. These problems are partly addressed in some recent works in reinforcement \citep{Flennerhag:2020wg} and supervised \citep{Almeida:2021} learning, which extend the standard white-box formulation by replacing learned hyper-parameters with learned functions of hand-picked \textit{context} features. Broadly speaking, context features can be any low-dimensional statistics of optimization, agent, or environment that carry information about suitable hyperparameter schedules. Examples of such context features in RL are reward histories, temporal difference errors and task beliefs. We refer to this approach as \emph{contextual meta-gradients}. While sharing some similarities with black-box meta-gradients, contextual meta-gradients differ in two important ways: (1) which update functions can be learned is more constrained for contextual meta-gradients (as the update rule can only be changed through hyperparameters) and (2) contextual meta-gradients have access to different kind of information (selected context features instead of entire history of gradients or parameters as is typical for black-box methods). 

In this paper, we hypothesize that: (1) the addition of context to meta-gradients is particularly well-suited for non-stationary environments with repeated structure, as it enables the optimizer to generalize from previously seen contexts and instantly pick good hyper-parameter values; (2) the advantage of all meta-gradients methods (with or without context) over well-tuned fixed hyperparameter schedules increases with the rate of environment non-stationarity, as different environment conditions may require very different learning hyperparameter values.

To examine the first hypothesis, we compare the performance of meta-gradients with and without contextual information across several environments. Since the addition of contextual information enables the meta-learner to learn hyperparameter schedules that generalize across learning contexts, we look at how increasing the context richness (i.e. amount of information given to the meta-learner) affects the performance. 
We find that making the hyperparameters a learned function of context features almost always helps training (Section \ref{subs:performance}), although some contexts are much more useful than others (Section \ref{subs:context}). We complement these findings by examining learned schedules and functions of context (Section \ref{subs:meta_functions}), where we show that the use of context is necessary for fast adaptation of hyperparameter values in response to the environment changes, and that learned functions of context are meaningful. 
Lastly, we investigate the adaptation ability of meta-gradients as the rate of environmental non-stationarity increases, with a particular interest in potential advantages of context in highly non-stationary environments (Section \ref{subs:nonstationarity}). We find that \emph{without} contextual information, meta-gradients provide small to inconsistent advantage over fixed hyperparameter values in highly non-stationary environments. On the other hand, meta-gradients \emph{with} contextual information provide a much more consistent advantage. 

 \vspace{-0.1cm}

%% file: core/method.tex
\section{Background} \label{sec:mg}

We follow the meta-gradient literature and make a distinction between the parameters $\param$ (e.g. weights of the policy and value networks) and the meta-parameters $\mparam$ (e.g. learning rates, discounts, weights of regularization losses). The parameters $\param$ are trained by minimizing an inner loss function $\loss^{\inne}_\eta(\param, \mathcal{D})$ parameterized by the meta parameters $\mparam$, whereas the meta-parameters $\mparam$ are trained to minimize the outer loss  $\loss^{\oute}(\mparam, \mathcal{D})$, where in both cases $\mathcal{D}$ refers to the rollout data used to estimate the loss. The parameters and meta-parameters are generally optimized at two different time-scales; we denote by $i$ the inner loop and by $k$ the outer loop iteration step. 

Most works in RL rely on entropy-regularized actor-critic algorithms \citep{Xu:2018metagradient, Zahavy:2020se}, where the inner loss objective $\loss^{\inne}(\param, \mathcal{D})$ consists of the following three terms (policy, value and entropy loss respectively):

\begin{equation}
\loss^{\inne}(\param, \mathcal{D}) = \alpha_{\pi} \loss_{\pi}(\param, \mathcal{D}) + \alpha_{v} \loss_{v}(\param, \mathcal{D}) + \alpha_{\ent} \loss_{\ent}(\param, \mathcal{D}) \label{eq:inner}    
\end{equation}

%

The hyperparameters required for computation of each of the three terms (for example, discounts $\gamma$ and bootstrapping parameters $\lambda$ in $n$-step returns) or the weights of individual losses, can become tunable meta-parameters. The outer loss objective generally consists of the same three terms, weighted by hyperparameters $\alpha^{\oute}_{\pi}, \alpha^{\oute}_{v}, \alpha^{\oute}_{ent}$: 

\begin{equation}
\loss^{\oute}(\mparam, \mathcal{D}) = \alpha^{\oute}_{\pi} \loss_{\pi}(\param(\mparam), \mathcal{D}) + \alpha^{\oute}_{v} \loss_{v}(\param(\mparam), \mathcal{D}) + \alpha^{\oute}_{\ent} \loss_{\ent}(\param(\mparam), \mathcal{D}), \label{eq:outer}    
\end{equation}

where the data $\mathcal{D}$ used for estimating the outer loss, could be coming from a separate rollout or the same data as used in the inner loss. The outer loss objective depends on the meta-parameters $\mparam$ through their influence on the learned parameters $\param$. This dependence is given by the update rule: $ \param_{i+1} = f(\param_i, \mparam_k, \mathcal{D}_{i})$, where $\mathcal{D}_{i}$ refers to the data used to compute the losses at $i$-th iteration. For example, for stochastic gradient descent, $\param_{i+1} = \param_i - \text{lr} \nabla_\param \loss^{\inne}_\eta(\param_i, \mathcal{D}_{i}),$ which is parameterized by $\mparam_k$ through $\loss^{\inne}_\eta(\param_i, \mathcal{D}_{i})$. Since backpropagating through the entire history of updates is too computationally demanding and may not provide a good gradient estimate due to non-stationarity intrinsic to RL, the meta-gradients are truncated to the last $K$ inner loop updates, which we refer to in Section \ref{sec:exps} as \emph{meta rollout length}. 

\section{Contextual Meta-Gradients} \label{sec:smg}

Contextual meta-gradients \citep{Flennerhag:2021bml, Almeida:2021} extend white-box meta-gradients by replacing the meta-parameter variables $\mparam_k$ in the update rule with parameterized functions $\mparam = g_{\omega_k}(\cdot)$ of context features $c_i$: $f(\param_i, g_{\omega_k}(c_i), \mathcal{D}_{i})$. The parameters $\omega_k$ of the meta-parameter function $g_{\omega}(\cdot)$ are learned by optimizing the outer loss, which is now $\loss^{\oute}(\omega, \mathcal{D})$, with the index $k$ referring to the value at outer optimization step $k$.
The meta-parameter function $g_{\omega}(\cdot)$ at $i$-th inner update takes context features $c_i$ as inputs, and outputs the value of the meta-parameter $\mparam$ used in the $i$-th inner update. For example, $g_{\omega}(\cdot)$ could be a neural network that takes average TD error over $i$-th batch as input and predicts the coefficient of $L_2$ regularization used in the $i$-th inner update. Also note that depending on how the context features are constructed and shared, the changes introduced by contextual meta-gradients allow for different meta-parameter predictions between individual inner updates or even individual samples contributing to the inner loss.


The context features can be any learning or environment statistics that capture information relevant for optimal meta-parameter schedules. Since contextual meta-gradients involve a function approximation in learning the meta-parameter function $g_{\omega}(\cdot)$, in practice, their usefulness relies on this meta-network being able to generalize with respect to context. For instance, when the learning process exhibits a cyclical pattern throughout the lifetime, contextual meta-gradients can improve upon standard meta-gradients as changes in learning dynamics are predictable and meta-network can learn the association between context and meta-parameter values. For example, if the context indicates there has been a drop in agent's performance, the meta-network $g_{\omega}(\cdot)$ predicting the rate of exploration can learn to associate this drop with an increase in the rate of exploration. 


In our experiments, the context features were chosen to indicate changes in task at hand and the agent performance (e.g. statistics of rewards, TD errors, values), all of which may require different meta-parameter values. Lastly, as in previous work with contextual meta-gradients \citep{Flennerhag:2021bml, Almeida:2021}, the gradients do not propagate through the context features.

\section{Non-Stationary Environments} \label{sec:envs}

To study non-stationarity, we chose to focus on environments with discrete and regular changes in reward and transition function. The regularity of changes enables control over the degree of non-stationarity by increasing or decreasing the length of the interval between two changes (Section \ref{subs:nonstationarity}) and it makes the learned meta-parameter schedules more interpretable (Section \ref{subs:meta_functions}). The agent interacts with these environments via a single-stream of experience as in the standard RL formulation (there is only one environment and not multiple copies of it, as is common in distributed RL frameworks). What follows is a description of two such environments that we use in our experiments, with more information and visualizations available in Appendix \ref{sec:appendix_envs}.



\textbf{Two Colors.} We start with a gridworld environment introduced in \cite{Flennerhag:2021bml}, where the agent is tasked with picking up one of the two items. Picking up one of the items results in reward +1 or -1, after which the agent and two objects are randomly re-spawned. Which object carries positive reward changes every 100,000 steps. The agents used in our experiments are memory-less, hence the optimal policy has to be re-learned after each task switch, which requires increasing the rate of exploration after the task switches. 

\textbf{Switching MDPs.} In our second non-stationary setting, the agent interacts with a sequence of grid worlds, where a new grid world is sampled every 100,000 steps from a set of $N$ grid worlds. The reward function and transition function of each grid world are randomly generated. 

We use this environment to study how changes in the environment dynamics changes, in addition to the reward function, affect meta-gradient methods. Additionally, in contrast to the Two Colors environment, two consecutive tasks may not be as different from each other and contain some positive transfer. This implies that in some cases, the optimal exploration strategy upon a task change may not be to explore too much. We experimented with two variants of Switching MDPs with different number of grid-worlds $N$. In the first case $N$=4 MDPs so that the agent is repeatedly exposed to the same tasks, and can improve its learning over time. In the second case, $N$=1000 MDPs, and the agent is very unlikely to experience the same task twice, requiring constantly learning almost from scratch in each MDP. This case is interesting, because we can study the generalization of contextual meta-gradients to unseen contexts. As tasks effectively do not repeat, distribution of context features which are given to the meta-parameter function after each task switch is much more irregular. For example, we don't observe regular periodic drops and increases in mean reward as we do in Two Colors environment.  

%% file: core/related_work.tex
\section{Related Work} \label{sec:related_work}

The focus of our paper is on white-box methods for learning adaptive optimizers \citep{Bengio:2000hypergradients, Maclaurin:2015hypergradients}. These methods tune the meta-parameters (i.e. tunable subset of hyper-parameters) of the learning update by computing the gradient of the outer loss with respect to meta-parameters, where the outer loss depends on meta-parameters both directly and through the history of last K parameter updates. In RL, this approach has been used to tune various optimization hyper-parameters, including discount and bootstrap parameters \citep{Xu:2018metagradient}, off-policy corrections \citep{Zahavy:2020se}, auxiliary rewards and tasks \citep{Zheng:2018le, veeriah2021discovery} and weights of rewards in return estimates \citep{Wang2019BeyondED}.

Among these methods, \cite{Flennerhag:2021bml} and \citet{Almeida:2021} propose learning meta-parameters as functions of context features, with the application in reinforcement and supervised learning respectively. While the focus of \cite{Flennerhag:2021bml} is on reducing myopia and conditioning problems of meta-gradients by developing an improved outer loss, in their experiments on non-stationary environments, they parameterize meta-parameters as a function of contextual features (in their case, reward statistics). However, they do not study the importance of including contextual information or provide comparison to alternative contextual features. In \citet{Almeida:2021}, learning meta-parameters as function of features is the primary focus. In contrast to most prior work in white-box meta-gradients, \citet{Almeida:2021} formulate optimization of meta-parameters as a reinforcement learning problem. The meta-parameter function is a policy modifying the current values of corresponding meta-parameters, trained on a distribution of source tasks with a designed reward function. The context features are selected to facilitate transfer of learned optimizers to the target tasks. In contrast to our work, their focus is on achieving high performance on supervised learning tasks instead of analysis of meta-gradients in non-stationary RL environments.


Alternatively, black-box methods parametrize the entire update rule as a neural network, and learn it from scratch by training on a distribution of supervised \citep{Andrychowicz:2016tf} or reinforcement learning \citep{Kirsch:2019im, Oh:2020discoveringrl, Kirsch2021IntroducingST} tasks. In RL, most of these methods require training over multiple tasks or lifetimes, which makes them not directly applicable to our setting of interest. The most relevant work is \cite{xu2020meta}, which develops a black-box method that trains an optimizer over a single lifetime. In black-box methods, the inputs to the learned optimizer functions are typically a history of parameters and gradients, but more similar to context features explored in our work, they can also include the entire rollout trajectories \citep{xu2020meta, Oh:2020discoveringrl}. However, these works do not explore the importance of selecting inputs for the learned optimizer and its effects on training in non-stationary environments.

%% file: core/experiments.tex
\section{Experiments} \label{sec:exps}

We designed the experiments with the goal of answering the following questions about the behaviour of meta-gradients in non-stationary environments: 

\textbf{Do meta-gradients benefit from contextual information?} We hypothesise that ability to learn meta-parameter schedules as functions of contextual information will enable faster adaptation in non-stationary environments, as the optimizer can leverage knowledge from previously seen contexts and instantly utilize good meta-parameter values without needing to slowly tune them. We compare the performance of contextual meta-gradients and baselines in Section \ref{subs:performance}.

\textbf{What functions of context are learned in this process?} To explain the performance difference between methods, we examine the meta-parameter schedules during training and the learned meta-parameter functions of context in Section \ref{subs:meta_functions}.
    
\textbf{What should the contextual information be?} The choice of context features could be essential for learning schedules that generalize over the training. 
Here we look into increasingly rich contexts: what is the effect of adding particular candidate contexts and can too much information be detrimental for generalization? We shed light on these questions in Section \ref{subs:context}. 

\textbf{How does the performance of meta-gradients depend on the rate of non-stationarity?} As the rate of change of the environment increases, we hypothesized that the ability to adapt the meta-parameters becomes more important. Furthermore, the addition of context to meta-gradients should lead to further advantages, as the they enable faster adaptation. The advantage of meta-gradients under different rates of non-stationarity is examined in Section \ref{subs:nonstationarity}.

\subsection*{Experimental Axis}

To ensure conclusions we make are robust, we run the experiments along several axis: (i) agents in inner loop, (ii) outer loop objectives, (iii) context features:

\textbf{Agents.} We use two different kinds of RL agents: Actor-Critic (AC) \citep{Sutton2000ac} and $Q(\lambda)$ \citep{Peng:1994qlambda}. The parameters of an AC agent are updated every 16 environment steps, and those of $Q(\lambda)$ agent at each step. If using AC agent, we tune the coefficient of entropy loss, whereas if using $Q(\lambda)$ agent, we tune the $\epsilon$ parameter of $\epsilon$-greedy exploration. The details of the agent architectures can be found in Appendix \ref{sec:appendix_hypers}.
    
\textbf{Outer Loop Objectives.} In experiments combining meta-gradients (MG) with AC agents, the outer loss is a sum of policy loss $\loss_{\pi}$ and a target entropy loss $\loss_{\ent}$ (the weight of this entropy loss is a fixed hyperparameter $\alpha^{(outer)}_{\ent}$):

\begin{equation}
    \loss_{\mg}^{\oute} = \loss_{\pi} + \alpha^{\oute}_{\ent} \loss_{\ent}.
\end{equation}

If using Bootstrapped Meta-Gradients (BMG) \citep{Flennerhag:2021bml}, the outer loss is KL divergence between the target policy $\pi_{\hat{\param}}$ and $K$-step bootstrap $\pi_{\param_K}$:
\begin{equation}
    \loss_{\bmg}^{\oute} = KL(\pi_{\hat{\param}} || \pi_{\param_K}),
\end{equation}

where the target policy is the policy reached after $K+L-1$ steps of inner loop optimization. For a longer description of BMG objective and it can be made differentiable with respect to $\epsilon$ when used with $Q(\lambda)$ agents, see Appendix \ref{sec:appendix_bmg}.

\textbf{Context Features.} We compare two kinds of contextual features, simple features referred to as \emph{Reward} (only using reward statistics) and more complex features referred to as \emph{Rich} (using reward, TD error and value statistics).
If the context is \emph{Reward}, each context feature is either a history of the last $H$ observed rewards ($Q(\lambda)$ agent) or a history of mean rewards observed in each of the last $H$ rollouts (AC agent). If the context is \emph{Rich}, we compute the same history for reward, TD error and values. In AC experiments, rich context also includes a history of last $H$ standard deviations in addition to means. We use $H=10$ in experiments with AC agents, and $H=100$ in experiments with $Q(\lambda)$ agents. Each context feature is normalized and scaled to be in the [-1, 1] range, before being concatenated and passed down to the learned meta-parameter function. More detailed description of context features used in our experiments can be found in Appendix \ref{sec:appendix_context}.

In each experiment, \textbf{we report only the results for the best hyper-parameter setting}. For the baselines (which do not adapt the meta-parameters), we include the meta-parameter of interest into the hyperparameter sweep. For MG objectives, we sweep over the meta learning rate, meta rollout length ($K$) and target entropy loss coefficient ($\alpha^{\oute}_{\ent}$). For BMG methods, we sweep over meta learning rate, meta rollout length ($K$) and target rollout length ($L$). For fairness, the size of hyperparameter sweeps is the same for all meta-gradient methods. For more details on the hyper-parameter sweeps and values, see Appendix \ref{sec:appendix_hypers}.

\begin{table}[t]
    \caption{Two Colors with AC (\textbf{left}) and $Q(\lambda)$ Agents (\textbf{right}). 
        Mean and standard deviation of total return after 10M steps (10 seeds). 
        \textbf{Left:} Only meta-gradient methods \emph{with} context features significantly outperform the AC baseline. 
        \textbf{Right:} All meta-gradient methods significantly outperform the $Q(\lambda)$ baseline, with the highest performance obtained by meta-gradients with Reward context features.
    }
    \begin{minipage}[t]{0.55\linewidth}
        \centering
        \as{1}
        \begin{tabular}{l|ccc}
            & \multicolumn{3}{c}{\bf Context Features} \\ 
            \hline
            {\bf Method} & None & Reward & Rich \\
            \hline
            AC              &  1.24 (0.09) & N/A & N/A  \\
            \hdashline
            AC-MG          &  1.29 (0.08) & 1.58 (0.06) & 1.62 (0.10) \\
            AC-BMG         &  1.32 (0.08) & 1.74 (0.03) & \textbf{1.79 (0.07)} \\
            \hline
        \end{tabular}
    \end{minipage}
    \hspace{1em}
    \begin{minipage}[t]{0.4\linewidth}
    \centering
        \begin{tabular}{l|cc}
            & \multicolumn{2}{c}{\bf Context Features} \\ 
            \hline
            {\bf Method} & None & Reward \\
            \hline
            Q($\lambda$)              & 0.58(0.07) & N/A \\
            \hdashline
            Q($\lambda$)-BMG         & 1.78(0.03) & \textbf{1.93(0.03)} \\
            \hline
        \end{tabular}
    \end{minipage}
    \label{tab:two_colors}
\end{table}

\subsection{Do Meta-Gradients Benefit from Adding Contextual Information?} \label{subs:performance}

We start the experiments by testing how adding contextual information to meta-gradients effects training. If our hypothesis is true, we expect to see the advantages of adding contextual information across different agents and outer loop objectives. Results in bold indicate the best mean performance among the compared methods.

\begin{table}[t]
\as{1}
    \caption{
        Switching MDPs with $Q(\lambda)$ Agent. 
        Mean and standard deviation of total return after 20M steps (10 seeds). Contextual information was necessary to gain a significant improvement over the baseline with meta-gradients. The improvement is greater when the number of different MDPs (i.e. variety of tasks experienced during a lifetime) is greater.
    }       
    \begin{center}
        \begin{tabular}{ll|cc}
        & & \multicolumn{2}{c}{\bf Number of MDPs} \\
        \hline
        {\bf Method} & & 4 & 1000 \\
        \hline
        $Q(\lambda)$                  & & 14.77 (1.78) & 12.03 (0.96) \\
        \hdashline
        $Q(\lambda)$-BMG             & & 14.79 (1.56) & 11.71 (0.71) \\
        $Q(\lambda)$-BMG-Reward      & & 15.00 (2.21) & 13.36 (0.68) \\
        $Q(\lambda)$-BMG-Rich        & & \textbf{16.03 (1.40)} & \textbf{13.63 (0.64)} \\
        
        \hline
        \end{tabular}
    \label{tab:switching_mdp_ql}
    \end{center}
\vspace{-0.5cm}

\end{table}


\textbf{Two Colors: Actor Critic Agent.} We first look at the AC agents trained on Two Colors (described in Section \ref{sec:envs}). In Table \ref{tab:two_colors} (left), we report the mean and standard deviation of the total reward after 10M environment steps. 

We find that all meta-gradient methods outperform the baseline with fixed meta-parameters (AC) and consistent with reports in \cite{Flennerhag:2021bml}, BMG performs better across methods. More importantly, the addition of context (see columns with Context Features set to Reward or Rich) is crucial for obtaining significant improvement over the baseline, with the richer features performing slightly better. 


\textbf{Two Colors: $Q(\lambda)$ Agent.} Next, we look at the $Q(\lambda)$ agents on Two Colors. In Table \ref{tab:two_colors} (right), we again report the mean and standard deviation of total returns after 10M environment steps. We tune the $\epsilon$ parameter of $\epsilon$-greedy exploration with only BMG objective, as the updated value function in this case is not a differentiable function of $\epsilon$ and consequently can not be straightforwardly optimized with regular MG objectives. 

We find that meta-gradients significantly outperform the non-adaptive baseline and the addition of context further boosts the performance. 


\textbf{Switching MDPs: $Q(\lambda)$ Agent}. Next, we look at $Q(\lambda)$ agent trained on Switching MDPs. In Table \ref{tab:switching_mdp_ql}, we report the mean and standard deviation of total rewards after 20M environment steps. We report only the experiments with $Q(\lambda)$ agents since AC agents were performing very poorly on this environment. Again, we tune only the  $\epsilon$ parameter. For these experiments, we also vary the number of different MDPs available in the environment (see Section \ref{sec:envs} on meaning and importance of these environment variations). 


We compare the results under the two environment variants of interest (4 and 1000 MDPs) in Table \ref{tab:switching_mdp_ql}. The addition of context was necessary to obtain significant improvements with meta-gradients, with Rich context resulting in the biggest improvement. The improvement is also much more significant in the regime where the MDPs effectively do not repeat (1000 MDPs). Because the consecutive tasks will typically be more similar compared to Two Colors, we hypothesize there is less need to re-learn in the regime with a small number of tasks (4 MDPs), requiring less adaptation of meta-parameters. 

\subsection{What Meta-Parameter Schedules and Functions are Learned?} \label{subs:meta_functions}

In this section, we wish to examine the predictions of the learned meta-parameter functions throughout training. For the ease of analysis, we focus on BMG with Reward context features (BMG-Reward) while only tuning one meta-parameter. 

\begin{figure}[t]

    \centering
    \begin{subfigure}[b]{0.45\textwidth}
    \centering
        \includegraphics[width=\textwidth]{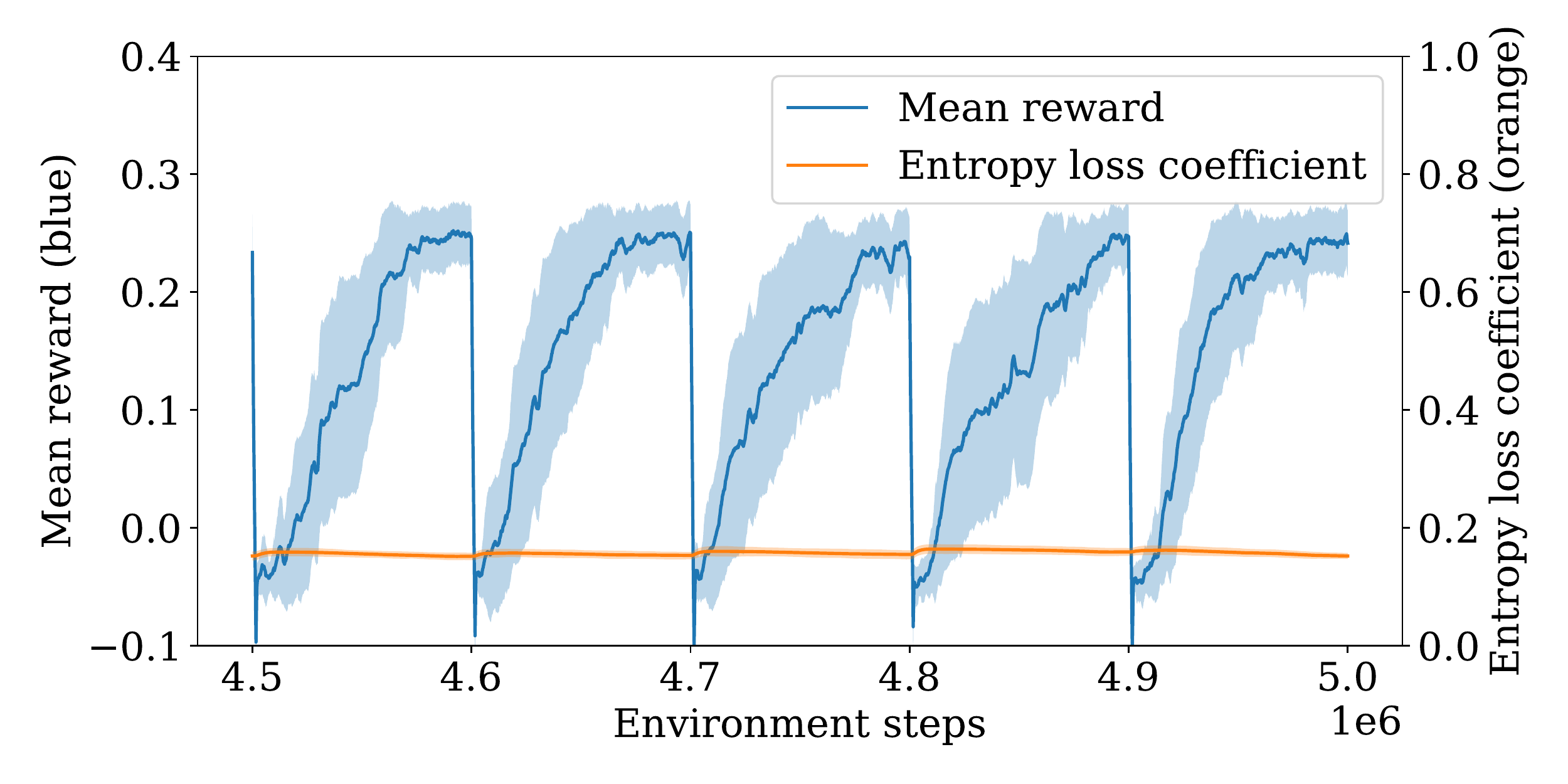}
        \caption{BMG.}
        \label{fig:mg_reward_vs_entropy}
    \end{subfigure}
    \begin{subfigure}[b]{0.45\textwidth}
    \centering
        \includegraphics[width=\textwidth]{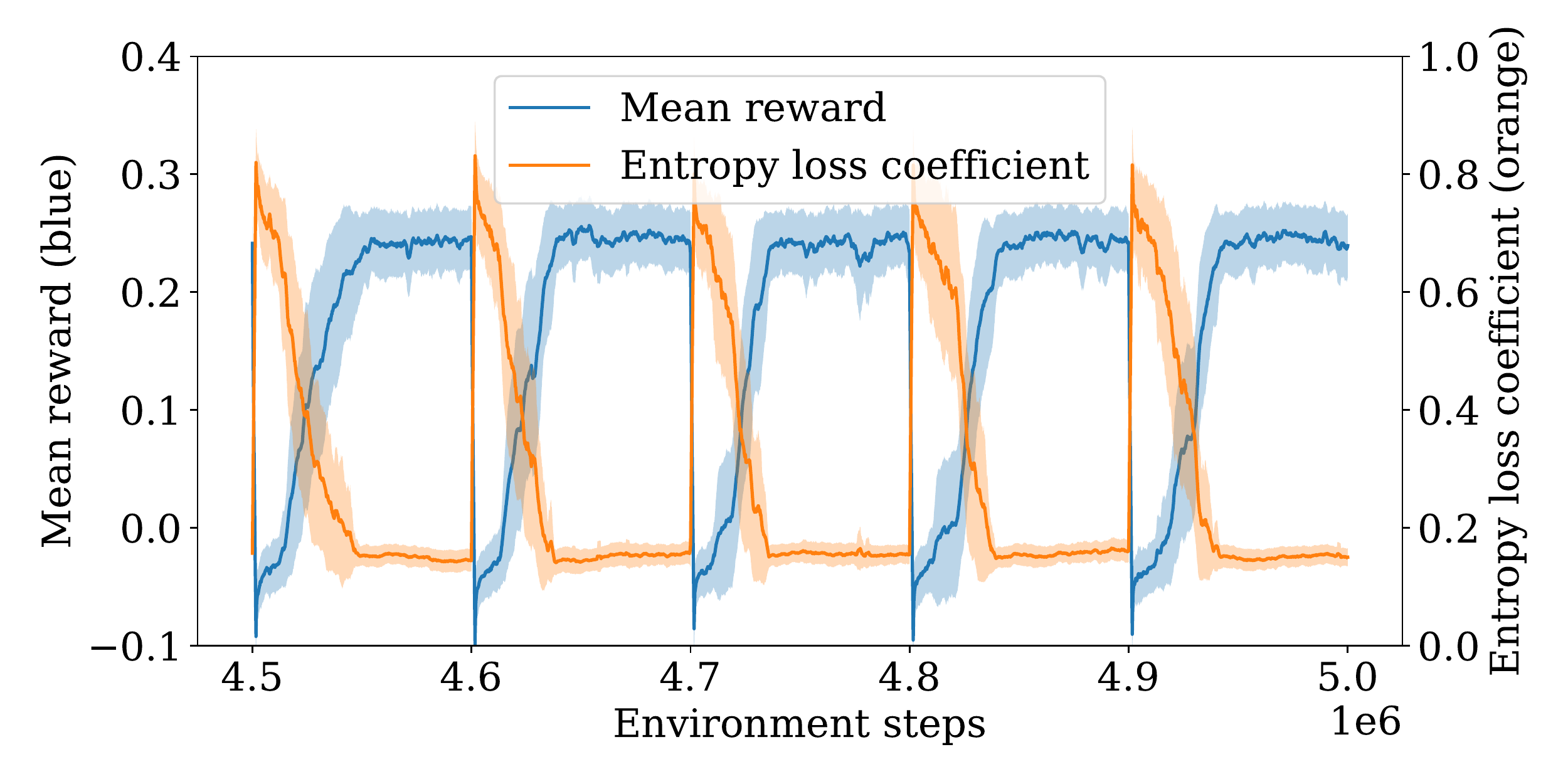}
        \caption{BMG-Reward.}
        \label{fig:smg_reward_vs_entropy}
    \end{subfigure}
    \caption{
        Entropy coefficient (orange) vs. mean reward (blue) during training for: (a) AC with BMG, (b) AC with BMG and reward context. 
        We report values at each timestep in the middle of training averaged over 10 seeds.
        The error margins represent 95\% confidence intervals.
        The drops in mean reward (blue) at regular correspond follow the task switches. 
        Without access to contextual information, meta-gradients learn an almost constant entropy schedule. By adding reward context, learned entropy schedule strongly responds to drops in mean reward.  
    }    
\end{figure}

\textbf{Learned Schedules.} 
We look at the relationship between the learned meta-parameter schedule and the observed rewards for AC agents with BMG trained on Two Colors. In Figure \ref{fig:mg_reward_vs_entropy}, meta-gradients do not rely on context features, and in Figure \ref{fig:smg_reward_vs_entropy}, meta-gradients utilize Reward context features. The two curves in both figures are: predicted entropy loss coefficient during training (orange curve) and mean reward over rollout (blue curve). 

When meta-gradients do not have access to reward context (Figure \ref{fig:mg_reward_vs_entropy}), the mean reward takes longer to recover after the drop following each task switch. Furthermore, entropy coefficient is almost constant. 
Note that this can not be explained by too low meta learning rate: the results shown here are under the best hyper-parameter configuration and the best performing meta learning rates were never the highest values in the sweep. In contrast, when the reward context is added (Figure \ref{fig:smg_reward_vs_entropy}), the entropy coefficient rapidly increases after each task switch to allow the agent to explore. As a result, the mean rewards of the agent are better as we observed in the previous section, suggesting that the addition of context is crucial for obtaining fast adaptation.

\textbf{Learned Functions.} Next, we look at how the context to meta-parameter mapping itself changes during a lifetime. Our methodology is inspired by visualization of model trajectories in \cite{erhan2010does}. Due to low dimensionality of inputs and outputs of meta-parameter function, we can select a small number of representative context inputs and for each, track the corresponding meta-parameter prediction during training. Note that we do not train on these context inputs, we just log the output for each while the meta-parameter function is trained as usual.  

\begin{figure}[b]
    \centering
    \begin{subfigure}[b]{0.45\textwidth}
    \centering
        \includegraphics[width=\textwidth]{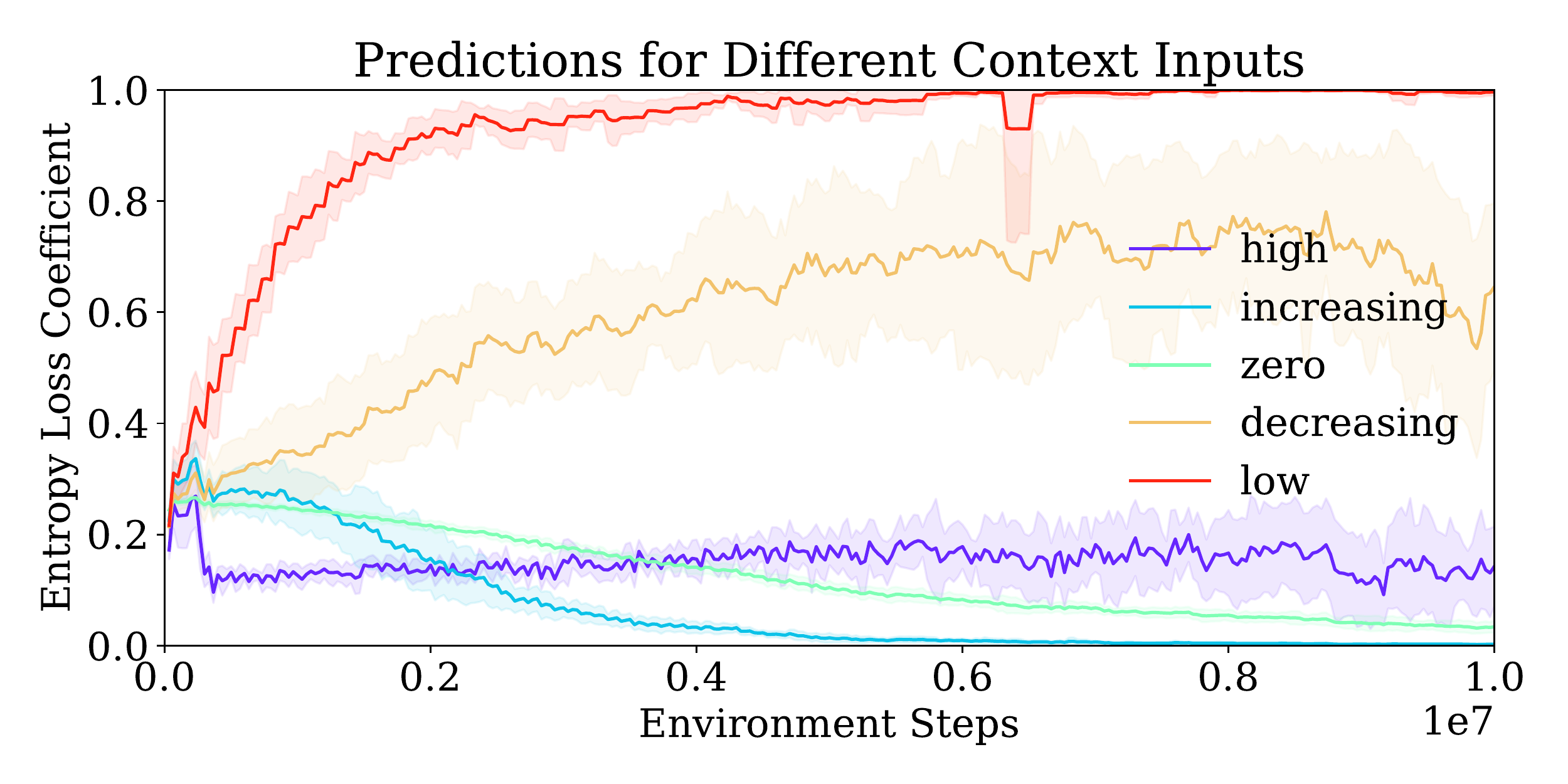}
        \caption{Two Colors with AC-BMG-Reward.}
        \label{fig:probes_two_colors_reward}
    \end{subfigure}
    \begin{subfigure}[b]{0.45\textwidth}
    \centering
        \includegraphics[width=\textwidth]{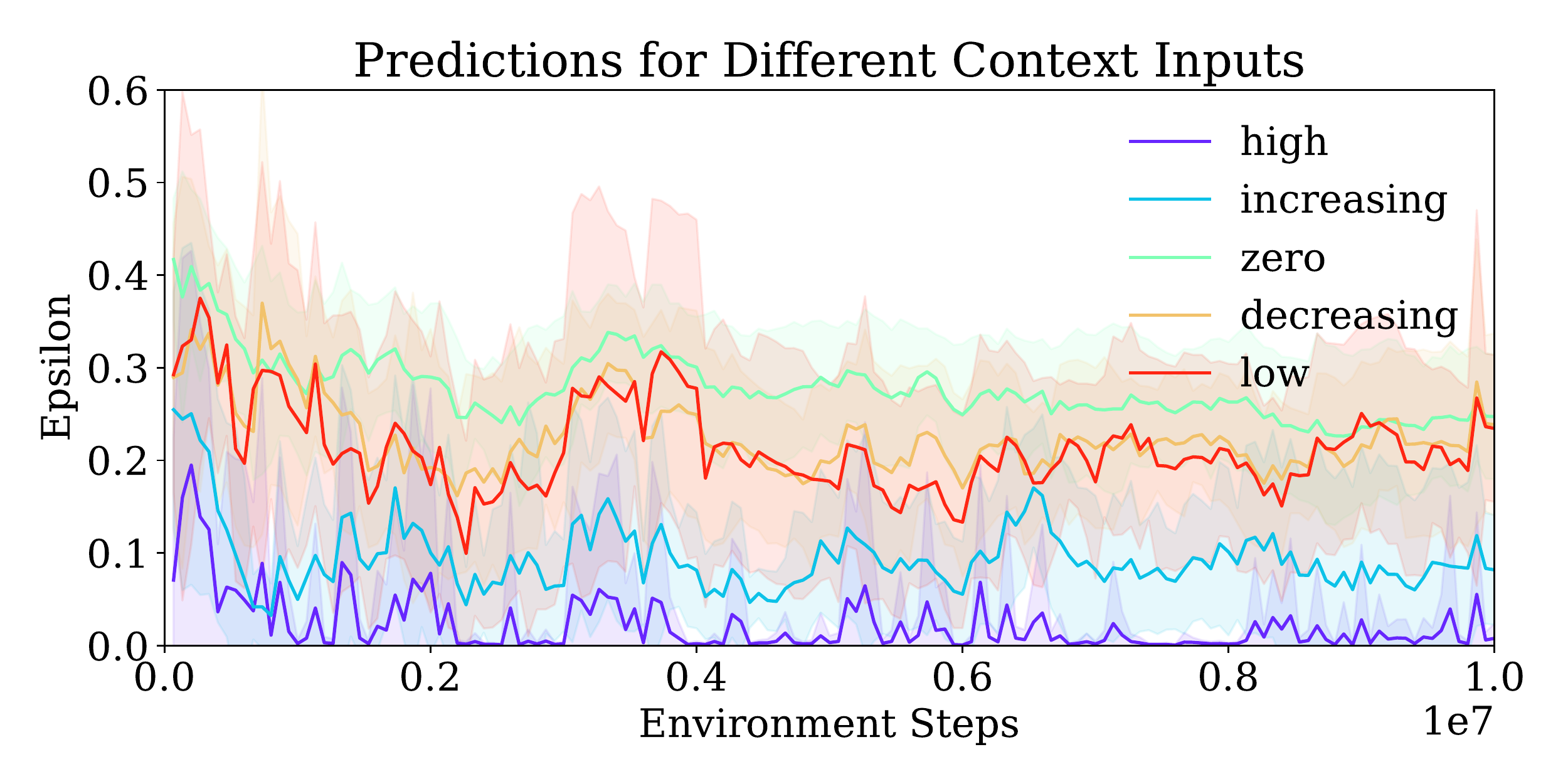}
        \caption{Switching MDPs (1k) with $Q(\lambda)$-BMG-Reward}
        \label{fig:probes_switch_mdp_reward}
    \end{subfigure}
    \caption{
    Predicted values of exploration meta-parameters for five qualitatively different reward context features as inputs to the meta-parameter function, as measured during training in: 
    (a) Two Colors with AC-BMG-Reward, (b) Switching MDPs (1000 MDPs) with $Q(\lambda)$-BMG-Reward.
    Each curve is averaged over 10 random seeds with the error margins representing one standard deviation.
    In (a), the $\alpha_{\text{ent}}$ is predicted when the rewards are the lowest, and the lowest when the rewards are increasing. In (b), the highest $\epsilon$ is predicted for lowest or zero rewards, and the lowest for highest rewards.
    }    
    \vspace{-0.5cm}
\end{figure}

The five context inputs were selected to represent qualitatively different learning situations: constantly high context values (\emph{"high"}), monotonically increasing from lowest to highest value (\emph{"increasing"}), constant zero (\emph{"zero"}), monotonically decreasing between the highest and lowest value (\emph{"decreasing"}) and constantly low context (\emph{"low"}). For example, for rewards bounded in $[-1, 1]$ and $H=3,$ the context inputs are: $\text{"high"}=[1, 1, 1]$,  $\text{"increasing"}=[-1, 0, 1]$,  $\text{"zero"}=[0, 0, 0]$,  $\text{"decreasing"}=[1, 0, -1]$ and  $\text{"low"}=[-1, -1, -1]$. In our experiment, $H=10$ and the range is $[-1, 1]$ due to pre-processing of context features. Note that some of the context inputs have the same mean (e.g., \emph{"increasing"}, \emph{"zero"} and \emph{"decreasing"}  all have mean zero), however they capture qualitatively different behaviors. For example, the \emph{"decreasing"} input corresponds to a context likely observed just after switching the task, while the \emph{"increasing"} probe is likely observed as the agent starts improving in a new task.

We visualize the learned meta-functions of AC agent trained with BMG-Reward on Two Colors in Figure \ref{fig:probes_two_colors_reward}, and those of $Q(\lambda)$ agent trained with BMG-Reward on Switching MDPs (1000 MDPs variant) in Figure \ref{fig:probes_switch_mdp_reward}. First, we can observe that different inputs results in very different meta-parameter predictions and that the meta-parameter functions seem to converge during training (the small local changes are likely due to non-stationary training distribution), hence addition of context enables convergence instead of just tracking as for vanilla MG. 
Next, we look at the values of predicted meta-parameters for each of the five context inputs. We can observe that the differences in predictions are sensible: in Figure \ref{fig:probes_two_colors_reward}, the entropy loss coefficient is highest when the rewards are at low or decreasing, and the lowest when the rewards are high or decreasing; whereas in Figure \ref{fig:probes_switch_mdp_reward}, the exploration is highest when the rewards are low, decreasing or zero and lowest when the rewards are high. The predictions are different for context inputs with the same mean value ("increasing", "decreasing", "zero"), indicating learned function responds to more complex patterns than just mean values of features. Lastly we note that for Switching MDPs, we see more variability in learned functions across different training seeds. This is likely due to a more complex interaction between encountered tasks and stochasticity in generating and sampling tasks (i.e. how much transfer there is between consecutive tasks will depend on which two tasks are sampled).

\begin{figure}[t]
    \centering
    \begin{subfigure}[b]{0.45\textwidth}
    \centering
        \includegraphics[width=\textwidth]{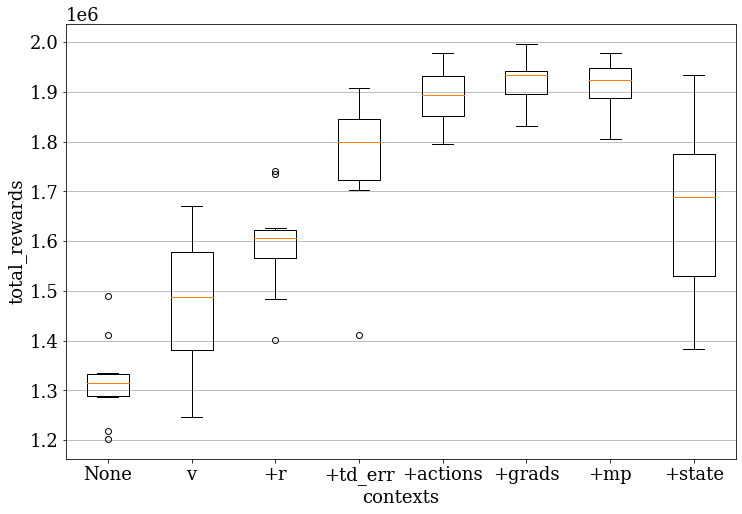}
        \caption{AC-BMG ($\alpha_{\ent}$)}
        \label{fig:gray_box_two_colors}
    \end{subfigure}
    \begin{subfigure}[b]{0.45\textwidth}
    \centering
        \includegraphics[width=\textwidth]{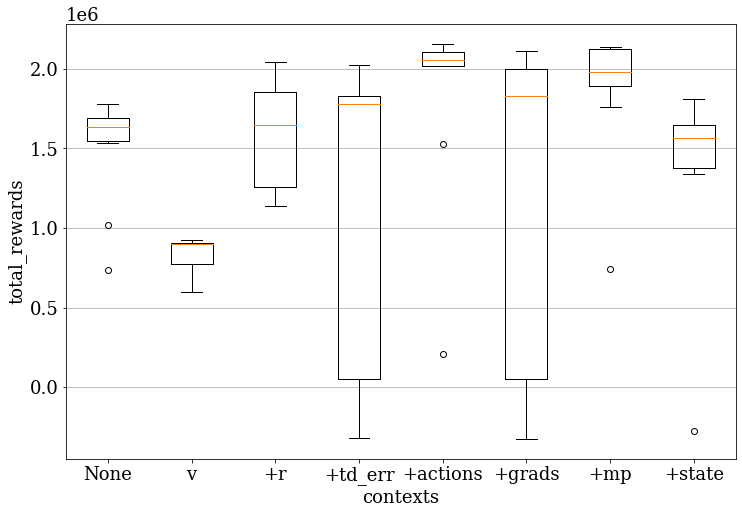}
        \caption{AC-BMG ($\alpha_{\ent}$. $\alpha_{L2}$)}
        \label{fig:gray_box_two_colors_l2}
    \end{subfigure}
    \caption{
    Performance of meta-gradients as a function of increased context richness: (a) AC-BMG with tuned entropy loss coefficient, (b) AC-BMG with tuned entropy and $L_2$ loss coefficients. 
    Total reward at 10M environment steps (10 seeds). 
    Starting without contextual information (None), in each column, from left to right, we add the statistics of the following quantities as features to the context: value (v), reward (+r), TD error (+td\_err), action probabilities (+actions), cosine distance between gradients (+grads), previous values of meta-parameters (+mp) and state visitation (+state).
    With some exceptions, adding more information leads to increase in mean performance. 
    }
\end{figure}

\subsection{How Important is the Choice of Context?} \label{subs:context}

In the following set of experiments, we inspect how making the context more rich effects performance. For this goal, we focus on Two Colors with the AC agent and BMG objective. We consider two variants; one that tunes the entropy coefficient (Figure \ref{fig:gray_box_two_colors}) and one that tunes the coefficients of both the entropy and the $L_2$ loss (Figure \ref{fig:gray_box_two_colors_l2}). 
The x-axis of each figure corresponds to the degree of richness of the context. On the left, we start with no context, and then add one-by-one the statistics of following quantities as context features: value, reward, TD error, action probabilities, cosine distance between last two gradients, past meta-parameter predictions and state visitation statistics. For each of these quantities (except cosine distance and past meta-parameter predictions), the features are a history of means and standard deviations calculated over the last $H$ rollouts. For cosine distance and past meta-parameter predictions, the features are just a history of their last $H$ values. Since the number of features is large, we decrease the history length $H$ from 10 to 4. If we were to use $H=10$, the dimension of richest context input would be 660, which could make the optimization of meta-function too challenging (see Appendix \ref{sec:appendix_context} for dimensions of each feature).  

The only cases where context features harm the performance are inclusion of statue visitation features in Figures \ref{fig:gray_box_two_colors} and \ref{fig:gray_box_two_colors_l2}, and relying on just value features in Figure \ref{fig:gray_box_two_colors_l2}. 
In the case of adding state visitation features, the performance drop is likely due to over-fitting. State features have much higher dimensionality compared to other features, yet they are not informative of the agent performance and hence good meta-parameter values. We did not find evidence of over-fitting for other features. Generally speaking, richer features led to better performance as meta-parameter function has more signal to leverage and can learn to rely on features that carry information about good meta-parameter values while ignoring the others. Note that the features that seem to help the most (reward, TD error), are strongly correlated with agent performance.


\begin{figure}[t]
    \centering
    \begin{subfigure}[b]{0.45\textwidth}
    \centering
        \includegraphics[width=\textwidth, height=5cm]{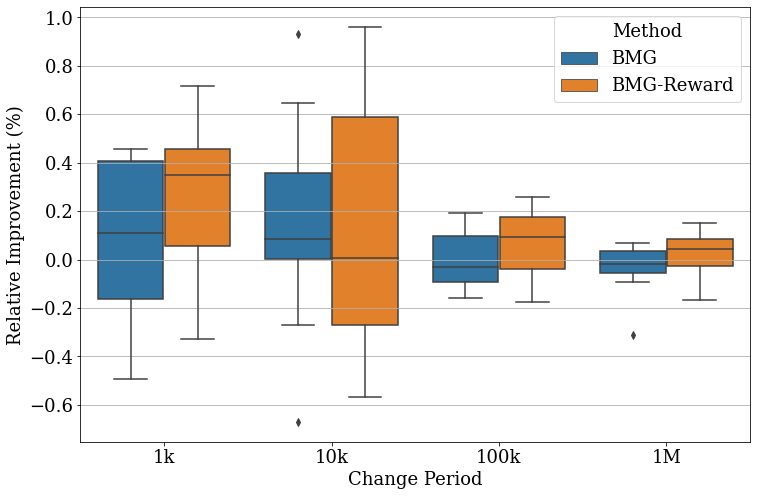} 
        \caption{4 MDPs}
        \label{fig:switching_mdp_rel_freq_4}
    \end{subfigure}
    \begin{subfigure}[b]{0.45\textwidth}
    \centering
        \includegraphics[width=\textwidth]{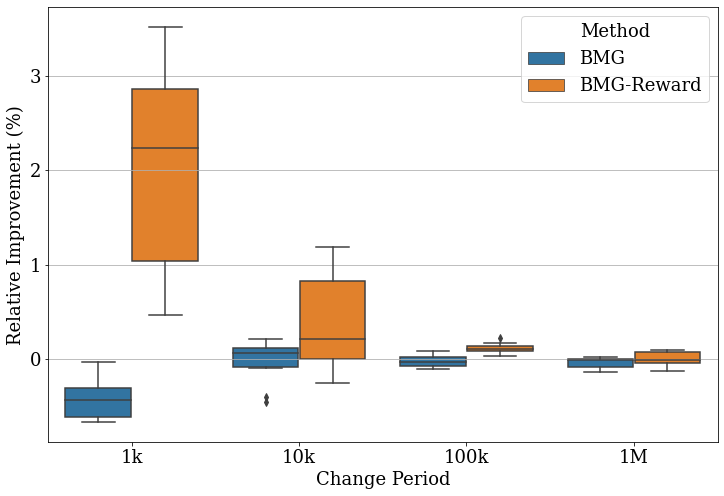}
        \caption{1000 MDPs}
        \label{fig:switching_mdp_rel_freq_1000}
    \end{subfigure}
    \caption{Relative improvement with meta-gradients over the $Q(\lambda)$ baseline under different rates of environment non-stationarity: (a) for Switching MDPs with 4 MDPs, (b) Switching MDPs with 1000 MDPs. 
    Meta-gradients with Reward context provide a more reliable performance boost, with the biggest improvement when the rate of non-stationarity and the number of different tasks are high.}    
\end{figure}

\subsection{How do Meta-Gradients Perform under Different Rates of Non-Stationarity?} \label{subs:nonstationarity}

Lastly we look at how the degree of non-stationarity in the environment effects the performance of meta-gradients. We control the degree of non-stationarity by changing how often the tasks switch -- shorter change periods correspond to higher rates of non-stationarity. In all of the Figures included in this section, the medians and interquartile ranges are computed over 10 random seeds.

\begin{wrapfigure}{r}{0.45\textwidth}
    \centering
    \includegraphics[width=0.45\textwidth]{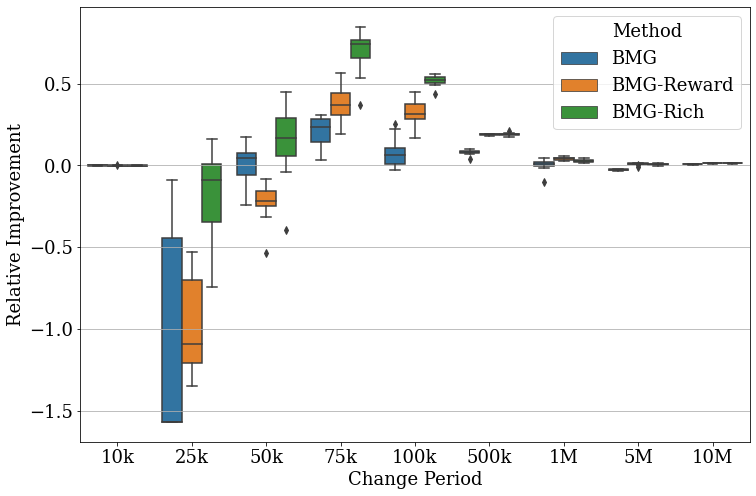}
    \caption{Two Colors: Relative improvement over the fixed meta-parameter AC baseline and under different rates of non-stationarity. 
    The regime in which meta-gradients provide a significant advantage over the baseline is the greatest for meta-gradients with rich context [50k, 500k]. All meta-gradient fail to beat the baseline when the rate of non-stationarity is too high (25k). 
    }
    \label{fig:two_colors_rel_freq}
    \vspace{-0.5cm}
\end{wrapfigure}

\textbf{Two Colors.} In this experiment, we compare the relative improvement of different meta-gradient methods over the AC baseline for various non-stationarity rates (Figure \ref{fig:two_colors_rel_freq}). The relative improvement is defined as percent increase or decrease in performance (as measured in total reward after 10M steps) over the mean performance of the baseline (with the same non-stationarity rate). The absolute (non-relative) values of all the methods can be found in the Appendix \ref{sec:appendix_freq}. We use BMG as an outer loop objective and tune entropy rate coefficient. The compared methods are: BMG (no context), BMG-Reward and BMG-Rich. As before, we report only the results for the best hyper-parameter configurations.

We find that when the environment changes too rapidly, at around 50,000 steps, the performance of meta-gradients deteriorates. 
We expected to find that improvement brought on by meta-gradients is greater when the environment changes more rapidly. This expectation was met up to a point: BMG with a rich context were less affected in addition to outperforming context-less meta-gradients for any rate of change, indicating that providing more information can provide enough signal to enable meta-learning even in this regime.


\textbf{Switching MDPs.} Figure \ref{fig:switching_mdp_rel_freq_4} and Figure \ref{fig:switching_mdp_rel_freq_1000} present the relative improvement of meta-gradients over the $Q(\lambda)$ baseline in the Switching MDPs environment (for $N=4$ and $N=1000$ respectively). The absolute (non-relative) values can be found in the Appendix \ref{sec:appendix_freq}. 

We find that the use of context becomes more useful as we increase the rate of environment non-stationarity and the number of different tasks, where fast adaptation of meta-parameters becomes more relevant.

%% file: core/appendix.tex
54\section{Appendix} \label{sec:appendix}

\subsection{Bootstrapped Meta-Gradients} \label{sec:appendix_bmg}

As an alternative to outer loss described in Section \ref{sec:mg}, \cite{Flennerhag:2021bml} propose the following objective: the meta-parameters are trained to minimize a distance to a target which has been bootstrapped from the meta-learner. This method is referred to as Bootstrapped Meta-Gradients (BMG). In the best studied version of BMG, the meta-gradients are propagated through the last $K$ updates, while minimizing the KL-divergence between the policy parametrized with $\param_K$ and a bootstrap target $\pi_{\hat{\param}}$ obtained by optimizing the policy for another $L-1$ steps under the meta-learned update rule:

\begin{equation}
    \loss_{\bmg}^{\oute} = KL(\pi_{\hat{\param}} || \pi_{\param_K}), \label{eq:bmg}
\end{equation}

where the hyperparameter $L$ is referred to as bootstrap target length. The use of target which is $L-1$ steps ahead (without increasing the number of steps the meta-gradients are backpropagating through) reduces the myopia of standard MG objectives, while the use of KL divergence reduces ill-conditioning of outer loop objective. The resulting adaptations of meta-parameters encourage reaching the target policy in a smaller number of inner optimization steps. 

When tuning the exploration rate $\epsilon$ of $Q(\lambda)$ agents, we use the following implementation from \cite{Flennerhag:2021bml} to make the outer objective differentiable with respect to $\epsilon$. In \eqref{eq:bmg}, the stochastic bootstrap policy $\pi_{\param_K}(a|s)$ and the target policy $\pi_{\hat{\param}}(a|s)$ are defined as:

\begin{equation}
    \begin{split}
        \pi_{\param_K}(a|s) = 
        \begin{cases}
            1 - \epsilon + \frac{\epsilon}{|A|} & \text{if} \ a = \underset{a'}{\argmax} \   q_{\param_K}(s,a') \\
           \frac{\epsilon_k}{|A|} & \text{else.}
        \end{cases}
    \end{split}
    \quad
    \begin{split}
        \pi_{\hat{\param}}(a|s) = 
        \begin{cases}
            1 & \text{if} \ a = \underset{a'}{\argmax} \  q_{\hat{\param}}(s,a') \\
            0 & \text{else.}
        \end{cases}        
    \end{split}    
\end{equation}

, where $q_{\param_K}(s, a)$ is the learned value function, the parameters $\hat{\param}$ are once again obtained by taking another $L-1$ update steps, and $|A|$ is the number of different actions. The resulting objective does not require differentiation through the update-rule, hence as in \cite{Flennerhag:2021bml}, we use $K=0$.

\begin{figure}[t]
    \centering
    \begin{subfigure}[b]{0.3\textwidth}
    \centering
        \includegraphics[width=\textwidth]{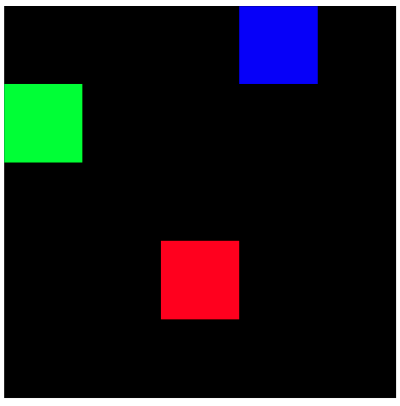}
        \caption{Two Colors.}
        \label{fig:two_colors}
    \end{subfigure}
    \begin{subfigure}[b]{0.5\textwidth}
    \centering
        \includegraphics[width=\textwidth]{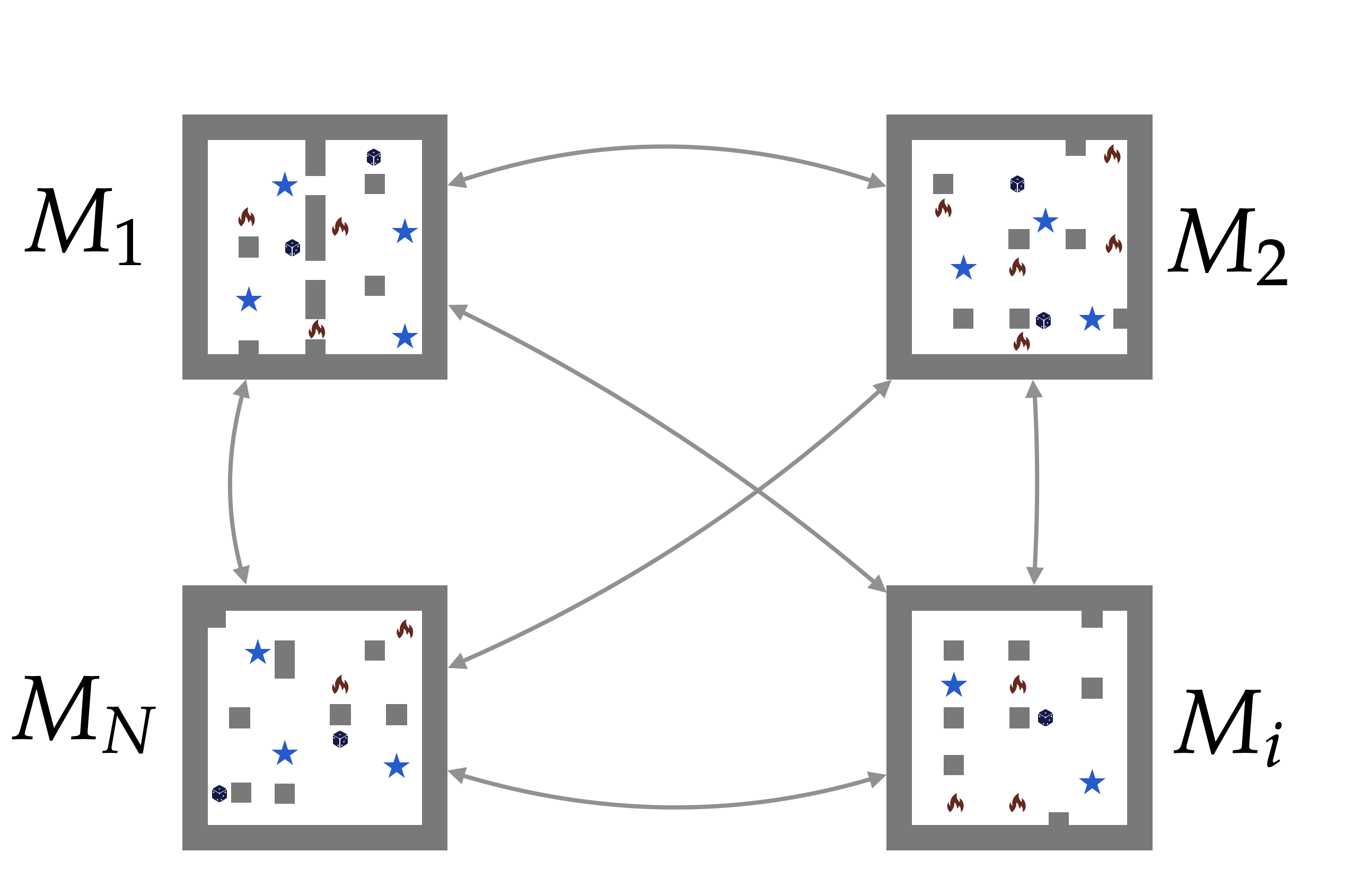}
        \caption{Switching MDPs.}
        \label{fig:switching_mdps}
    \end{subfigure}
    \caption{Illustration of Two Colors and Switching MDPs environments. (a) In Two Colors, the agent (green) is tasked with navigating to either blue or red square. (b) In Switching MDPs, the reward and transition function changes to one of the $N$ predefined randomly generated options at regular intervals.    
    }
\end{figure}

\subsection{Environments} \label{sec:appendix_envs}

In this section, we provide additional information about the environments used in this paper. The Two Colors environment is visualised in Figure \ref{fig:two_colors} and Switching MDPs in \ref{fig:switching_mdps}. Note that the range of reward in considered environments does not change during a lifetime, hence the context features related to reward (such as those based on reward, value and TD-error) are likely to stay within a certain distribution, which is important for generalization of context functions.

\subsubsection{Two Colors}

The dimension of the grid in Two Colors environment is $5 \times 5$. The observation space is constructed by concatenating one-hot encodings of $x$ and $y$ coordinates of positions of agents and two other objects. The total dimension of resulting observation space is $3 \times 2 \times 5 = 30$. 

\subsubsection{Switching MDPs}

The reward function for each MDP is generated in the following manner: for each state-action pair there is a 50\% chance of zero reward, 20\% chance of reward of +1, 20\% chance of reward -1, and 10\% chance of a random reward sampled uniformly from the interval [-1, 1]. The transition function for each MDP is a standard two dimensional grid world---states are characterized by an $(x,y)$ coordinate, and there are four actions that move the agent up, left, right, and down unless the intended cell is the edge of the grid or a wall is present. The transition function changes across the $N$ MDPs by the addition of a set of walls randomly placed throughout the grid. Up to 15 walls are placed per-MDP, in sequence, by randomly sampling unoccupied cells (by goal, agent, or another wall).

At the beginning of an experiment, a set of $N$ different MDPs is generated. At regular fixed intervals, the next grid world is sampled from this set uniformly with replacement. The set and the order of samples is uniquely determined by experiment random seed. If $N$ is large compared to the total number of task switches experienced during a lifetime (e.g. $N=1000$ with a change period $100,000$ and a lifetime of $20M$ steps), the probability of agent experiencing the same task multiple times is small. Note that since we measure the lifetime performance after $20M$ environment steps in our experiments, $N$ does not correspond to the number of different MDPs experienced during this lifetime.

\begin{table}[h]
        \caption{Hyper-parameters used in experiments with Actor-Critic agents. We denote in italic when a hyper-parameter is required in only some variants of the experiments (e.g. AC baseline, MG or BMG objective, contextual meta-gradients). 
        }
        \vspace{-0.3cm}

    \begin{center}
        \begin{tabular}{ll}
         Inner Learner: Actor Critic & \\
         \hline
         Optimizer & SGD \\ 
         Learning Rate & $0.1$ \\  
         Batch Size & 16 \\  
         $\alpha_{\ent}$ candidates \emph{(AC only)} & $[0, 0.1, 0.2, 0.4, 0.8]$\\    
         $\gamma$ & 0.99 \\     
         MLP hidden layers $(v, \pi)$ & 2 \\
         MLP feature size $(v, \pi)$ & 256 \\ 
         Activation Function & ReLU \\ 
        \end{tabular}
        \end{center}
        \begin{center}
        \begin{tabular}{ll}
         Meta-learner & \\
         \hline
         Optimizer & Adam \\ 
         $\epsilon$ (Adam) & $10^{-4}$ \\
         $\beta_1$, $\beta_2$ (Adam) & 0.9, 0.999\\
         Learning Rate candidates & $[10^{-3}, 10^{-4}, 10^{-5}, 10^{-6}]$ \\  
         $\alpha^{\oute}_{\ent}$ \emph{(MG only)} & $[0, 0.1]$ \\    
         $K$ candidates & $[1, 3, 6]$ \\
         $L$ candidates \emph{(BMG only)} & $[8, 16]$\\
         $H$ \emph{(contextual only)} & 10 \\
         MLP hidden layers \emph{(contextual only)} & 2 \\
         MLP feature size \emph{(contextual only)}) & 64 \\ 
         Activation Function \emph{(contextual only)} & ReLU \\ 
         Output Activation & Sigmoid \\ 
        \end{tabular}
        \label{tab:ac_hypers}
        \vspace{-0.28cm}
    \end{center}
\end{table}   

\begin{table}[h]
    \begin{center}
        \caption{Hyper-parameters used in experiments with $Q(\lambda)$ agents. We denote in italic when a hyper-parameter is required in only some variants of the experiments (e.g. $Q(\lambda)$ baseline, contextual meta-gradients). 
        }    
        \vspace{-0.1cm}

        \begin{tabular}{ll}
         Inner Learner: $Q(\lambda)$ & \\
         \hline
         Optimizer & Adam \\ 
         $\epsilon$ (Adam) & $10^{-4}$ \\
         $\beta_1$, $\beta_2$ (Adam) & 0.9, 0.999 \\
         Learning Rate candidates \emph{($Q(\lambda)$ only)} &  $[3 \cdot 10^{-3}, 10^{-4}, 3 \cdot 10^{-5}, 10^{-5}]$ \\  
         $\epsilon$ candidates \emph{($Q(\lambda)$ only)} & $[0.3, 0.1, 0.03, 0.01]$\\    
         $\lambda$ & 0.9 \\     
         $\gamma$ & 0.99 \\     
         MLP hidden layers $(q)$ & 2 \\
         MLP feature size $(q)$ & 256 \\ 
         Activation Function & ReLU \\ 
        \end{tabular}
        \end{center}
        
        \begin{center}
        \begin{tabular}{ll}
         Meta-learner & \\
         \hline
         Optimizer & Adam \\ 
         $\epsilon$ (Adam) & $10^{-4}$ \\
         $\beta_1$, $\beta_2$ (Adam) & 0.9, 0.999 \\         
         Learning Rate candidates \emph{(no context)} & $[10^{-2}, 3 \cdot 10^{-3}, 10^{-3}, 3 \cdot 10^{-4}]$ \\  
         Learning Rate candidates \emph{(with context)}& $[10^{-3}, 10^{-4}, 10^{-5}, 10^{-6}]$ \\  
         $L$ candidates & $[16, 32, 128]$\\
         $H$ \emph{(contextual only)} & 100 \\
         MLP hidden layers \emph{(contextual only)} & 2 \\
         MLP feature size \emph{(contextual only)}) & 128 \\ 
         Activation Function \emph{(contextual only)} & ReLU \\ 
         Output Activation & Sigmoid \\ 
        \end{tabular}
        \label{tab:ql_hyp}
    \end{center}
\end{table}

\newpage

\begin{table}[t]
    \caption{
        Summary of all context features used in experiments with AC agents. When description of $f_i^{(feat)}$ includes mean \& std, we are indicating that $f_i^{(feat)}$ is mean and standard deviation of the specified quantity, over the rollout data used to compute $i$-th update. Here we denoted the gradient with respect to inner loss at $i$-th update with $\nabla_{\param}\loss_i$. 
    }
    \begin{center}
        \begin{tabular}{lll}
         Feature Type & $f_i^{(feat)}$ & Feature Dimension\\
         \hline
          Reward & $r_t$ (mean \& std) & $2 \times H$ \\ 
         Value &  $v(s_t; \param_i)$ (mean \& std) & $2 \times H$ \\  
         TD Error & $r_t + \gamma v(s_{t+1}; \param_i) - v(s_t; \param_i)$ (mean \& std) & $2 \times H$ \\    
         Action Probabilities & $\pi(a|s_t; \param_i)$ (mean \& std) & $8 \times H$ \\     
         States & $s_t$ (mean \& std) & $60 \times H$ \\
         Cosine Distance Between Gradients & $1 - \frac{\nabla_{\param}\loss_{i-1} \nabla_{\param}\loss_{i-2}}{\| \nabla_{\param}\loss_{i-1} \| \|\nabla_{\param}\loss_{i-2} \|}$ & $H$ \\
         Meta-Parameters & $\mparam_{i-1}$ & $H$ \\
        \end{tabular}
    \label{tab:ac_context_features}
    \end{center}
\end{table}   
\begin{table}[t]
    \caption{
       Summary of all context features used in experiments with $Q(\lambda)$ agents. 
    }
    \begin{center}
        \begin{tabular}{lll}
         Feature Type & $f_i^{(feat)}$ & Feature Dimension\\
         \hline
          Reward & $r_t$ & $H$ \\ 
          Value &  $q(a_t, s_t; \param_i)$ & $H$ \\  
          TD Error & $r_t + \gamma \max_a q(s_{t+1}, a; \param_i) - q(s_t, a_t; \param_i)$ & $H$ 
        \end{tabular}
    \label{tab:ql_context_features}
    \end{center}
\end{table}  

\subsection{Experimental Setup and Hyper-parameters} \label{sec:appendix_hypers}

In this section, we provide further details on the experimental setup and hyper-parameters of agents and meta-learners used in Section \ref{sec:exps}. 

The hyper-parameters used in the experiments with AC agents are described in Table \ref{tab:ac_hypers}. The softmax policy and the value function are implemented by two separate feed-forward MLPs. The parameters are updated every 16 environment steps: given fixed parameters, the agent interacts with the environment for 16 steps collecting observations, rewards and actions into a rollout, which is then used to compute the inner loss. The inner loss consists of the following four terms (policy, value, entropy and $L_2$ loss respectively):

\begin{equation}
\loss^{\inne}(\param, \mathcal{D}) = \loss_{\pi}(\param, \mathcal{D}) +
\loss_{v}(\param, \mathcal{D}) + \alpha_{\ent} \loss_{\ent}(\param, \mathcal{D}) + \alpha_{\text{L2}} \loss_{\text{L2}}(\param) \end{equation}

In most of the experiments, the only meta-parameter is $\alpha_{\ent}$ and $\alpha_{L2}=0$ (i.e. there is no $L_2$ regularization). The exception is Section \ref{subs:context} (Figure \ref{fig:gray_box_two_colors_l2}), where we tune both $\alpha_{\ent}$ and $\alpha_{L2}$. 
In experiments with contextual meta-gradients, we use a feed-forward MLP $g_{\omega}(\cdot)$ that takes context features as inputs and predicts the meta-parameter value (i.e. $\alpha_{\ent} = g_{\omega}(c)$). When tuning two meta-parameters, we used two separate feed-forward MLPs, while taking the same context features as input (i.e. $\alpha_{\ent} = g_{\omega^{(\ent)}}(c)$ and $\alpha_{L2} = g_{\omega^{(L2)}}(c)$). The parameters $\omega$ of these meta-networks are trained by optimizing one of the two outer losses (see Section \ref{sec:exps}, Experimental Axis). The output of the meta-parameter network predicting $\alpha_{L2}$ has been scaled by a fixed quantity ($10^{-4}$) to prevent training instabilities caused by too strong forgetting. The weights of both policy and value networks were included in $\loss_{\text{L2}}(\param)$. 

The hyper-parameters used in the experiments with $Q(\lambda)$ agents are described in Table 4. The q-function is again a feed-forward MLP. The agent is optimized at each step, i.e. without batching. To avoid instabilities this could cause, we use a momentum term that maintains an exponentially moving average over gradients, with the discount factor $0.9$. We sweep over the learning rate of the inner learner only for the fixed meta-parameter baseline, for the meta-gradient methods, the inner learner's learning rate is set to $3 \cdot 10^{-5}$. The tuned meta-parameter is $\epsilon$ of the $\epsilon$-greedy exploration. The meta-parameter network is once again a feed-forward MLP, that takes context features as input and predicts $\epsilon$ (i.e. $\epsilon = g_{\omega}(c)$). The parameters $\omega$ of this MLP are trained by optimizing the outer loss, in this case a BMG objective described in Appendix \ref{sec:appendix_bmg}. 

In experiments with both AC and $Q(\lambda)$ agents, to ensure stable initial predictions, the meta-networks were pre-trained on random context inputs sampled from uniform distribution $[-1, 1]$, to predict an output in the middle of the possible meta-parameter range ($0.5$ for $\alpha_{\ent}$ and $\epsilon$, $5 \cdot 10^{-5}$ for $\alpha_{L2}$). Note that when using contextual meta-gradients, we do not sweep over the hyper-parameters introduced by the addition of context (such as the dimensions of meta-network and context history $H$), hence the size of the hyper-parameter sweep is the same for the meta-gradients with and without context.


\subsection{Context Features} \label{sec:appendix_context}

In this section, we describe in more details the construction of context features. 
At each update step $i$, the meta-parameters are computed by feeding context features $c_i$ into a meta-parameter function. The context features are a concatenation of one or several different types of features (e.g. reward, value, TD error), the type of features used in each experiments is described in the corresponding experiment's section. For example, when the context is specified as \emph{Reward}, the input to the meta-parameter function is $c_i = [c^{(reward)}_i]$, and when the context is \emph{Rich}, the input to the meta-parameter function is $c_i = [c^{(reward)}_i, c^{(value)}_i, c^{(td-error)}_i]$. For each different feature type (here denoted with "feat"), the features are a history of size $H$: 

\begin{equation}
c_i^{(feat)} = [\hat{f}_i^{(feat)}, \hat{f}_{i-1}^{(feat)}, ...,  \hat{f}_{i-H+1}^{(feat)}],    
\end{equation}

where $\hat{f}_i^{(feat)}$ is a quantity computed using statistics at $i$-th update step. Each of these quantities has been normalized (we used the statistics observed over a lifetime, but more generally, one could use a running average as an estimate) and passed through a $tanh(\cdot)$ function to ensure all features are in the [-1, 1] range. We will refer with $f_i^{(feat)}$ to quantities computed before this transformation.  
For example, in experiments described with AC-BMG-Reward, each $f_i^{(reward)}$ is a mean of rewards $r_t$ from the rollout $D_i$ which was used to compute $i$-th update. In experiments described with $Q(\lambda)$-BMG-Reward, because $i$-th update for $Q(\lambda)$ agents is calculated using data from only one environment step, $f_i^{(reward)}$ is the reward $r_t$ from only that step.   

The summary of all different feature types used in experiments with AC agents, including how  $f_i^{(feat)}$ is defined for that feature type and dimensions of that feature, can be be found in Table \ref{tab:ac_context_features}. Note though that when the feature is specified as \emph{Reward} (AC-MG-Reward and AC-BMG-Reward), we only used mean of rewards, and the corresponding feature dimension is $H$. The summary of all different feature types used in experiments with $Q(\lambda)$ agents can be found in Table \ref{tab:ql_context_features}.



\begin{wrapfigure}{r}{0.45\textwidth}
    \centering
    \includegraphics[width=0.45\textwidth]{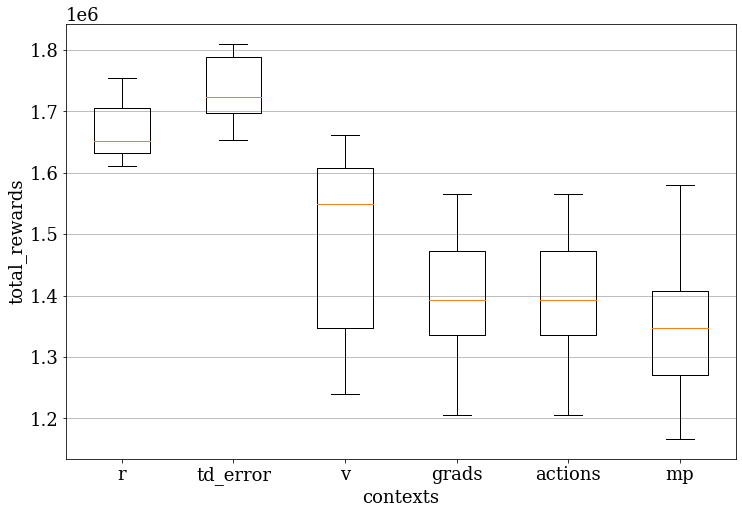}
    \caption{Two Colors: Comparison of performance using each of the contexts individually, measured in total return after 10M steps. The agent is Actor Critic, the outer loss is BMG and we tune entropy loss coefficient. The meaning of context labels is same as in Section \ref{subs:context}.
    The medians and interquartile ranges are computed over 10 random seeds.
    }
    \label{fig:two_colors_individual}    
\end{wrapfigure}

\subsection{Choice of Context Features: Individual Contexts} \label{sec:appendix_gray}
We supplement the results in Section \ref{subs:context}, Figure \ref{fig:gray_box_two_colors}, with a plot of context meta-gradients with each of the contexts individually (Figure \ref{fig:two_colors_individual}). We can observe that, looking at the each context individually, the highest gains are obtained with the contexts that highly correlate with the agent performance (reward, TD error, value).

\subsection{Different Rates of Non-Stationarity: Total Rewards} \label{sec:appendix_freq}

Lastly, we supplement the results in Section \ref{subs:nonstationarity} by illustrating how the performance of all methods drops as the rate of environment non-stationarity increases. 

\begin{figure}[b]
    \centering
    \includegraphics[width=0.75\textwidth]{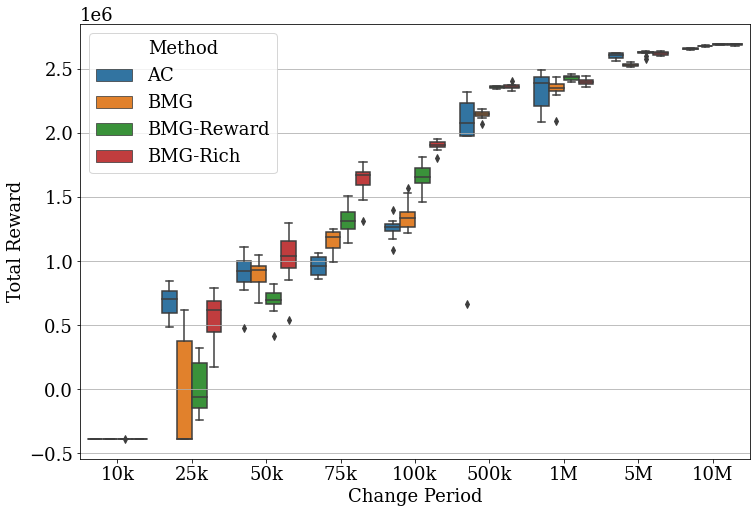}
    \caption{Two Colors: Comparison of methods under different rates of environment non-stationarity as measured in total return after 10M steps.
    The medians and interquartile ranges are computed over 10 random seeds.
    The regime in which meta-gradients provide a significant advantage over the baseline is the greatest for meta-gradients with rich context (50k-500k steps between task switches). All meta-gradient fail to beat the baseline when the rate of non-stationarity is too high (25k steps between task switches). 
    }
    \label{fig:two_colors_freq}    
\end{figure}

\begin{figure}[b]
    \centering
    \begin{subfigure}[b]{0.45\textwidth}
    \centering
        \includegraphics[width=\textwidth]{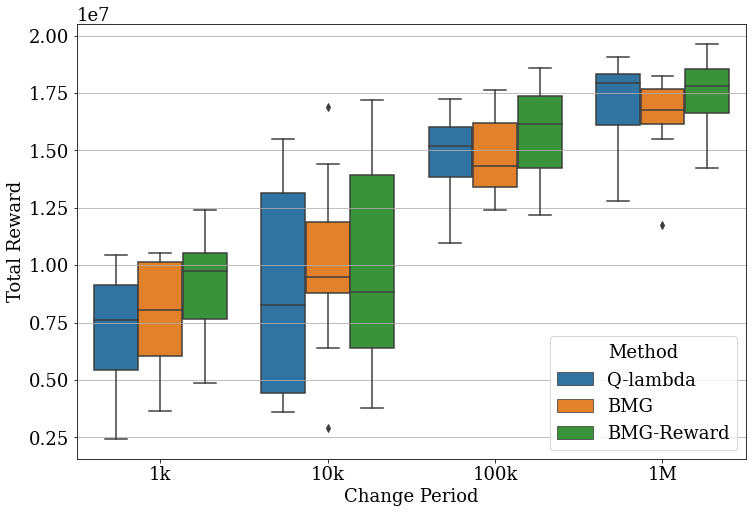}
        \caption{4 MDPs: Comparison of $Q(\lambda)$, $Q(\lambda)$-BMG and $Q(\lambda)$-BTMG-Reward.}
        \label{fig:switching_mdp_freq_4}
    \end{subfigure}
    \begin{subfigure}[b]{0.45\textwidth}
    \centering
        \includegraphics[width=\textwidth]{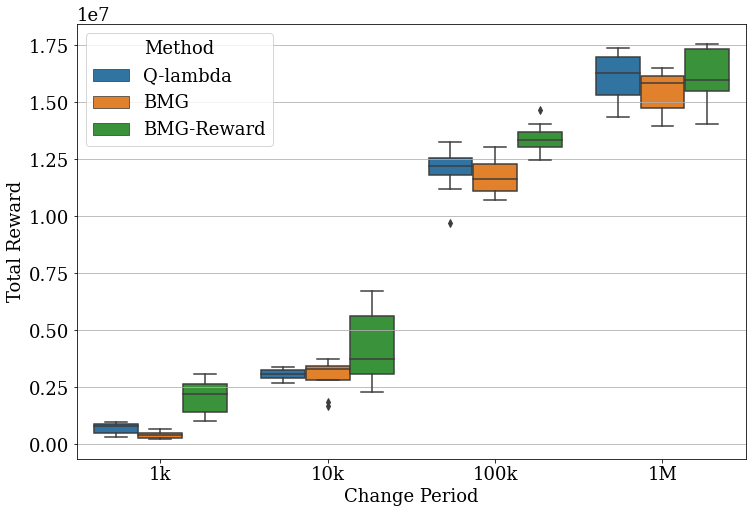}
        \caption{1000 MDPs: Comparison of comparison of $Q(\lambda)$, $Q(\lambda)$-BMG and $Q(\lambda)$-BTMG-Reward.}
        \label{fig:switching_mdp_freq_1000}
    \end{subfigure}
    \caption{Comparison of methods (measured in total return after 10M steps) under different rates of environment non-stationarity: (a) Switching MDPs Environment with 4 different MDPs, (b) Switching MDPs Environment with 1000 different MDPs. 
    The performance of all methods drops when the number of different tasks is very large and as the rate of non-stationarity increases. Meta-gradients with Reward context perform the best when the rate of non-stationarity is high and the number of different tasks are high. 
    }
\end{figure}

In Two Colors experiments (Figure  \ref{fig:two_colors_freq}), no learning occurs when the environment switches every 10k steps. In Switching MDPs, note that the performance drop is much more significant when the number of different MDPs is large (Figure \ref{fig:switching_mdp_freq_1000}), indicating that in this environment, the agents benefit from repeated exposure to the same MDP.

